%% file: paper.tex
\pdfoutput=1

\documentclass[11pt]{article}

\usepackage[final]{acl}

\usepackage{times}
\usepackage{latexsym}
\usepackage{enumitem}

\usepackage[T1]{fontenc}

\usepackage[utf8]{inputenc}

\usepackage{microtype}

\usepackage{inconsolata}

\usepackage{graphicx}

\usepackage{multirow}
\usepackage{adjustbox}
\usepackage{amssymb}
\title{Evaluating Small Language Models for News Summarization: Implications and Factors Influencing Performance}

\author{Borui Xu \\
  Shandong University \\
  \texttt{boruixu@mail.sdu.edu.cn} \\\And
  Yao Chen\thanks{Corresponding authors} \\
  National University of Singapore \\
  \texttt{yaochen@nus.edu.sg} \\
  \And
  Zeyi Wen \\
  HKUST(GZ) \& HKUST \\
  \texttt{wenzeyi@ust.hk}
  \AND
  Weiguo Liu\textsuperscript{*} \\
  Shandong University \\
  \texttt{weiguo.liu@sdu.edu.cn}
  \And
  Bingsheng He \\
  National University of Singapore \\
  \texttt{hebs@comp.nus.edu.sg}}

\begin{document}
\maketitle
\begin{abstract}


The increasing demand for efficient summarization tools in resource-constrained environments highlights the need for effective solutions. While large language models (LLMs) deliver superior summarization quality, their high computational resource requirements limit practical use applications. In contrast, small language models (SLMs) present a more accessible alternative, capable of real-time summarization on edge devices. However, their summarization capabilities and comparative performance against LLMs remain underexplored. This paper addresses this gap by presenting a comprehensive evaluation of 19 SLMs for news summarization across 2,000 news samples, focusing on relevance, coherence, factual consistency, and summary length. Our findings reveal significant variations in SLM performance, with top-performing models such as Phi3-Mini and Llama3.2-3B-Ins achieving results comparable to those of 70B LLMs while generating more concise summaries. Notably, SLMs are better suited for simple prompts, as overly complex prompts may lead to a decline in summary quality. Additionally, our analysis indicates that instruction tuning does not consistently enhance the news summarization capabilities of SLMs. This research not only contributes to the understanding of SLMs but also provides practical insights for researchers seeking efficient summarization solutions that balance performance and resource use.

\end{abstract}







\input{ARR_version/01_introduction_paper_debugger}
\input{ARR_version/02_barkground_paper_debugger}
\input{ARR_version/03methods}

\input{ARR_version/04experiment}

\section*{Acknowledgments}
This work is funded by the National Key R\&D Program of China (No. 2019YFA0709400), the NSFC Project (No. 62306256), the Guangzhou Science and Technology Development Projects (No. 2023A03J0143 and No. 2024A04J4458), and the Guangzhou Municipality Big Data Intelligence Key Lab (NO. 2023A03J0012). This work is also supported by the National Research Foundation, Singapore under its Industry Alignment Fund – Pre-positioning (IAF-PP) Funding Initiative. Any opinions, findings and conclusions or recommendations expressed in this material are those of the author(s) and do not reflect the views of National Research Foundation, Singapore. Borui's work was done when he was a visiting student at the National University of Singapore.

\bibliography{custom}

\appendix

\input{ARR_version/05appendix}

\end{document}

%% file: ARR_version/01_introduction_paper_debugger.tex
\section{Introduction}

Automatic text summarization is a fundamental challenge in Natural Language Processing (NLP) that plays a crucial role in various applications, including search engine optimization~\cite{2019soe}, financial forecasting~\cite{fintech}, and public sentiment analysis~\cite{twitter_news}. The ability to distill large volumes of information into concise summaries is particularly important in today's fast-paced digital environment, where users often seek quick insights from news articles. Recent advancements in large language models (LLMs) such as GPT-3~\cite{NEURIPS2020_gpt3}, OPT~\cite{opt}, and Llama2~\cite{touvron2023llama2openfoundation} have significantly enhanced the quality and coherence of generated summaries compared to traditional models~\cite{luhn1958automatic, dong-etal-2018-banditsum, zhang-etal-2018-neural, t5, genest-lapalme-2012-fully}. However, the hundreds of billions of parameters in these LLMs require substantial computational and storage resources. For example, deploying the Llama2-70B model at FP16 precision typically necessitates two Nvidia A100 GPUs with 80GB of memory each. This means that only organizations with significant computing power can offer LLM services, raising concerns about service stability and data privacy.

\begin{figure}
    \centering
    \includegraphics[width=1\linewidth]{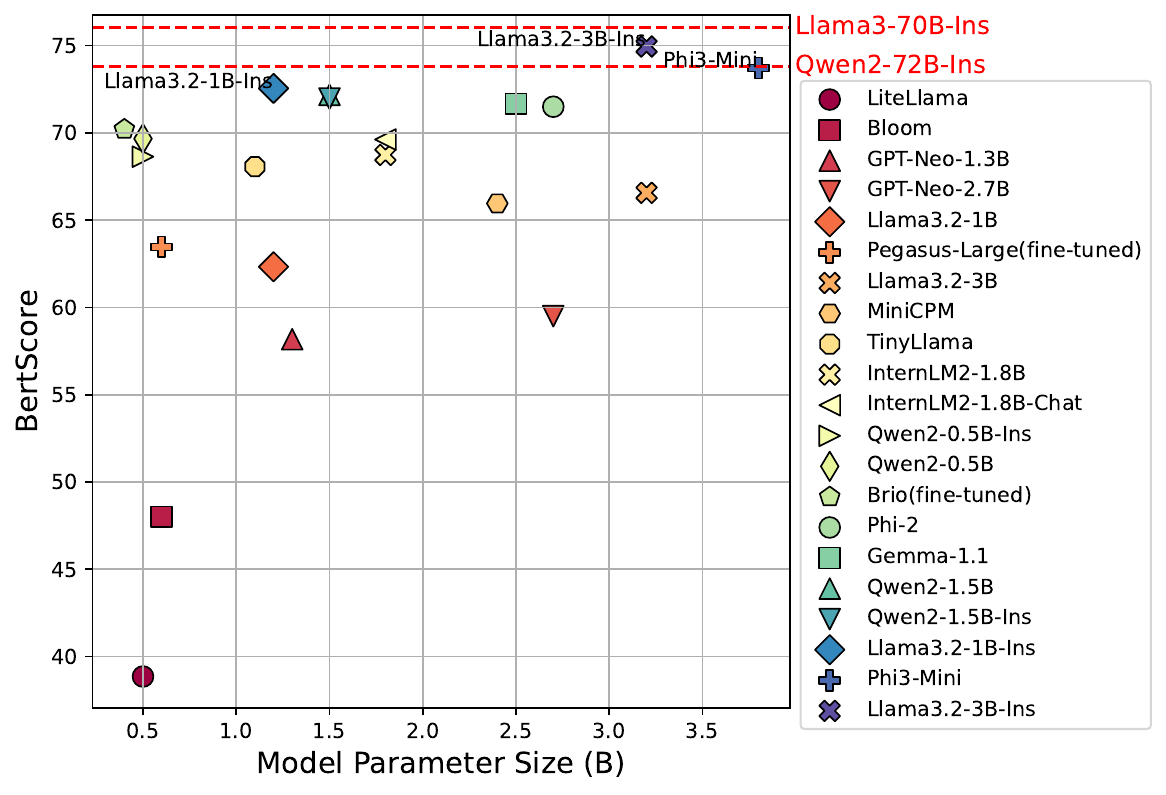}
    \caption{Variation of SLMs in size and BertScore for news summarization. }
    \label{fig:intro}
\end{figure}

As a result, general-purpose small language models (SLMs) have emerged as a potential alternative, as shown in Figure~\ref{fig:intro}. They share the same decoder-only architecture as LLMs but have fewer than 4B parameters~\footnote{Currently, there is no clear definition of SLMs, but they generally have fewer than 4 billion parameters.}~\cite{azure2024phi,qwen,abdin2024phi3,llama3.2}. Same as the LLMs, they generate specified content based on prompts. SLMs are designed to run efficiently on edge devices like smartphones and personal computers, making them more accessible. This accessibility makes SLMs an attractive option for applications requiring quick, stable, and low-cost solutions that also protect user privacy, such as summarizing news articles for mobile users or enhancing data processing privacy in sensitive environments.

Despite the promising capabilities of SLMs, their specific performance in the domain of news summarization remains underexplored. Previous studies, such as Tiny Titans~\cite{tiny_titan}, have evaluated specifically fine-tuned SLMs in meeting summarization. The adopted ROUGE metric~\cite{lin-2004-rouge} only reflects the quality in terms of relevance. In addition, broader evaluations of SLMs across various tasks have been evaluated but omitted news summarization~\cite{lu2024smalllanguagemodelssurvey}, highlighting the gap in understanding SLM in this area.

This paper aims to address these gaps by conducting a comprehensive evaluation of pre-trained general-purpose SLMs in news summarization, a process that abstracts news articles into shorter versions while preserving key information. We seek to answer the following research questions:

First, \textbf{How do different SLMs compare to each other, and how do they compare to LLMs?} We benchmark 19 SLMs on 2,000 news samples\footnote{The codes and results are available at \url{https://github.com/Xtra-Computing/SLM_Summary_Benchmark}}. Our findings reveal that the news summarization capabilities of different SLMs vary significantly as shown in Figure~\ref{fig:intro}. And the best-performing SLMs can generate news summaries comparable in quality to those produced by LLMs while producing shorter summaries. Notably, Phi3-Mini and Llama3.2-3B-Ins emerge as the top performers.

Second, \textbf{What is the impact of prompt complexity on the performance of SLMs?} We compare the impact of prompts with varying levels of detail on the quality of the summaries. Our findings indicate that prompt engineering has a limited impact on enhancing summarization abilities, with even complex prompts potentially degrading model performance. Thus, simple prompts suffice in practical applications.

Third, \textbf{How does instruction tuning affect the summarization capabilities of different SLMs?} We find that the effects of instruction tuning on SLMs vary significantly. While the Llama3.2 series models show substantial improvements after instruction tuning, which aligns with existing works~\cite{zhang2024benchmarking}, models like Qwen2 and InternLM2 exhibit minimal changes.


The rest of the paper is organized as follows: Section~\ref{sec:background} covers the background and related work, Section~\ref{sec:method} introduces reference summary generation, Section~\ref{sec:setup} illustrates the benchmark design, Section~\ref{sec:benchmark} presents presents the benchmarking results of SLMs, Section~\ref{sec:factor} analyzes the influencing factors on SLM summarization, and Section~\ref{sec:conclusion} concludes our findings.

%% file: ARR_version/02_barkground_paper_debugger.tex
\section{Background and Related Work}
\label{sec:background}

\subsection{Small Language Model}

Small language models (SLMs) are characterized by their reduced parameter count and computational requirements compared to larger LLMs. These models can be general-purpose or specialized, and this paper focuses on the evaluation of pre-trained general-purpose SLMs, which will be referred to simply as SLMs throughout the remainder of the paper. Structurally, SLMs and LLMs share a common architecture, consisting of stacked decoder-only layers from the transformer framework~\cite{attention_all_you_need}, generating outputs in an autoregressive manner. SLMs typically have fewer decoder layers, attention heads, and smaller hidden dimensions, and they are trained on smaller, high-quality datasets~\cite{lu2024smalllanguagemodelssurvey}. While there is no clear definition for the parameter count that qualifies a model as an SLM, models that can operate on consumer-grade devices are generally considered to fall within this category. This paper focuses on models smaller than 8GB in FP16 precision, corresponding to fewer than 4B parameters.

\subsection{Text Summarization and Its Evaluation}


\begin{figure}
    \centering
    \includegraphics[width=1\linewidth]{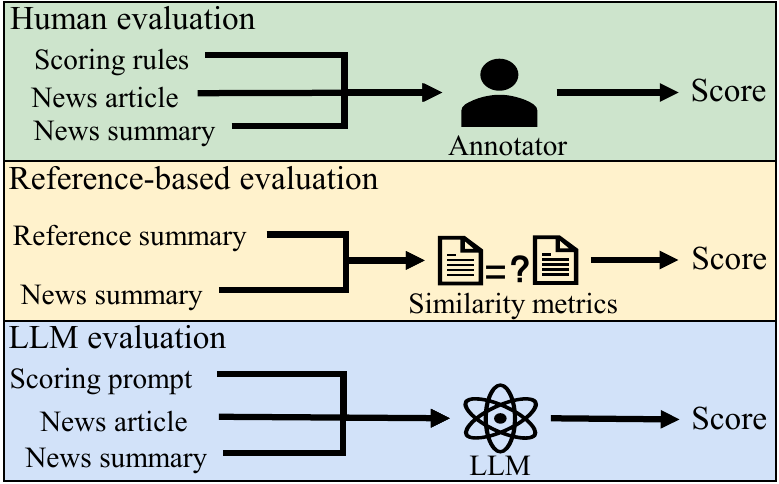}
    \caption{Comparison of text summarization evaluation. }
    \label{fig:evaluation_compare}
\end{figure}


Text summarization is the process of compressing a large volume of text into a shorter version while retaining its main ideas and essential information. Traditional text summarization models usually take a single news article as input and output a summary. However, SLMs require both a prompt and an article to generate a summary. 


The evaluation methods of text summarization can be categorized into three categories, as shown in Figure~\ref{fig:evaluation_compare}: human evaluation, reference-based evaluation, and LLM evaluation.

\noindent \textbf{Human evaluation} is an intuitive and effective method, involving the hiring of annotators to score the generated text. For instance, \citet{huang-etal-2020-achieved} employed human evaluation to evaluate 10 representative summarization models, revealing that extractive summary methods often performed better in human evaluation. The Summeval benchmark~\cite{fabbri2021summeval} utilized human evaluation methods to evaluate 23 traditional models and found Pegasus~\cite{pegasus} performed best. \citet{dead_summarization} used human evaluation to assess LLMs and found that they were more favored by evaluators. \citet{zhang2024benchmarking} also used human evaluation to evaluate the LLMs, discovering that instruction tuning had a greater impact on performance than model size. INSTRUSUM~\cite{liu2023benchmarking} evaluated instruction controllable summarization of LLMs by annotators and found them still unsatisfactory in summarizing text based on complex instructions.

\noindent \textbf{Reference-based evaluation} is a commonly used method. It scores generated text by calculating its similarity to reference texts using various metrics, and many such metrics have been proposed~\cite{lin-2004-rouge,papineni-etal-2002-bleu,sellam-etal-2020-bleurt,bert-score,bartscore}. In addition to human evaluation, \citet{huang-etal-2020-achieved} and ~\citet{fabbri2021summeval} also used many reference-based metrics to evaluated models and found metrics like Rouge and BertScore are very close to human evaluations. \citet{google-2023-benchmarking} used the Rouge metric to measure text similarity. Tiny Titans~\cite{tiny_titan} assessed the performance of small models in meeting summarization and found FLAN-T5 performed best.

\noindent \textbf{LLM evaluation} is explored by recent studies~\cite{goyal2022news,kendeer,gao2023human-like,liu2023benchmarking,wang2023chatgpt}. This approach involves guiding LLMs with prompts and providing both the summary and the original article for scoring. However, \citet{not-yet} indicated that LLMs may exhibit biases when evaluating different models, warranting further research to ensure fairness.

In conclusion, existing evaluations primarily evaluate the news summarization capabilities of traditional models or LLMs, leaving a gap in the evaluation of SLMs.

%% file: ARR_version/03methods.tex
\section{LLM-Augmented Reference-Based Evaluation}\label{sec:method}

As the reference-based method offers higher efficiency, better reproducibility, and lower cost compared to human and LLM evaluation~\cite{fabbri2021summeval,zhang2024benchmarking}, we use the reference-based method for large-scale SLM evaluation. However, some existing research~\cite{how_well,fabbri2021summeval} indicates that the poor quality of heuristic reference summaries in current datasets weakens their correlation with human preferences. Moreover, \citet{zhang2024benchmarking} find that using high-quality summaries as references can significantly improve alignment with human preferences. Thus, we use LLM-generated summaries as references in our SLM evaluation instead of relying on original dataset references. 

LLM-generated reference summaries offer two key advantages.
First, they have higher quality than original heuristic summaries; \citet{zhang2024benchmarking} find LLMs averaged 4.5 on a five-point scale, while original references scored only 3.6. Additionally, \citet{dead_summarization} find that LLM-generated summaries are preferred to those written by freelance authors by up to 84\% of the time. 
Second, LLMs produce high-quality summaries quickly, generating hundreds in a few hours, which benefits large-scale news evaluations. In contrast, a freelance writer typically takes about 15 minutes for a single summary~\cite{zhang2024benchmarking}. 

\begin{figure}
    \centering
    \includegraphics[width=1\linewidth]{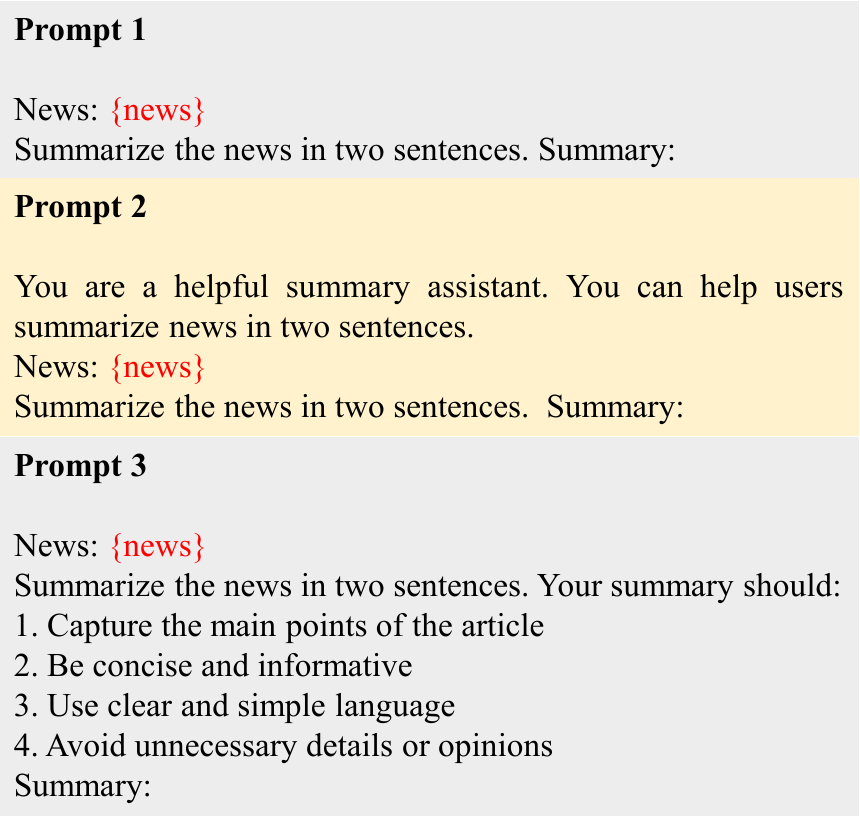}
    \caption{ The prompt templates for the language model to generate summaries. }
    \label{fig:prompt}
\end{figure}

We also verify whether the LLM-generated reference summaries can improve the effectiveness of reference-based evaluation on three human evaluation datasets~\cite{zhang2024benchmarking,fabbri2021summeval} in Table~\ref{tab:compare_kt_corr}. These datasets include summaries generated by various models as well as human ratings. We compare the correlation of three popular similarity metrics with human scoring using different references. BertScore and BLEURT calculate similarity based on fine-tuned language models, while RougeL measures similarity through textual overlap. We use prompt 2 in Figure~\ref{fig:prompt} as the input of the LLM. For all datasets, we consistently use two sentences to summarize the article. LLM-generated summaries significantly improve consistency than original references. BertScore performs best in both the relevance and coherence aspects.

\begin{table*}[]
\begin{center}
\resizebox{2\columnwidth}{!}{
\begin{tabular}{cccccccccc}
\hline
\multirow{2}{*}{Metric} &
  \multirow{2}{*}{\begin{tabular}[c]{@{}c@{}}Reference\\ type\end{tabular}} &
  \multicolumn{4}{c}{Relevance evaluation} &
  \multicolumn{4}{c}{Coherence evaluation} \\
 &
   &
  \begin{tabular}[c]{@{}c@{}}Bench\\ -CNN/DM\end{tabular} &
  \begin{tabular}[c]{@{}c@{}}Bench\\ -XSum\end{tabular} &
  SummEval &
  Average &
  \begin{tabular}[c]{@{}c@{}}Bench\\ -CNN/DM\end{tabular} &
  \begin{tabular}[c]{@{}c@{}}Bench\\ -XSum\end{tabular} &
  SummEval &
  Average \\ \hline
\multirow{3}{*}{RougeL}   & Original    & 0.6732 & 0.2647 & 0.3714 & 0.4364          & 0.4248 & 0.6765 & 0.1238 & 0.4084          \\
                           & Qwen1.5-72B & 0.7778 & 0.7059 & 0.4476 & \textbf{0.6438} & 0.5556 & 0.5000 & 0.2381 & \textbf{0.4312} \\
                           & llama2-70B  & 0.7778 & 0.6324 & 0.4095 & 0.6066          & 0.5294 & 0.3676 & 0.2000 & 0.3657          \\ \hline
\multirow{3}{*}{BertScore} & Original    & 0.6209 & 0.2206 & 0.4476 & 0.4297          & 0.4771 & 0.6912 & 0.4286 & 0.5323          \\
                           & Qwen1.5-72B & 0.7516 & 0.6324 & 0.8095 & 0.7312          & 0.5817 & 0.6029 & 0.6000 & \textbf{0.5949} \\
                           & llama2-70B  & 0.7516 & 0.6618 & 0.8286 & \textbf{0.7413} & 0.5294 & 0.6029 & 0.6190 & 0.5838          \\ \hline
\multirow{3}{*}{BLEURT}    & Original    & 0.5425 & 0.2647 & 0.181  & 0.3294          & 0.3203 & 0.6176 & 0.2381 & 0.3929          \\
                           & Qwen1.5-72B & 0.5817 & 0.7206 & 0.4476 & 0.5833          & 0.4902 & 0.4853 & 0.5429 & 0.5061          \\
                           & llama2-70B  & 0.5556 & 0.7059 & 0.5429 & \textbf{0.6015} & 0.5425 & 0.4706 & 0.6762 & \textbf{0.5631} \\ \hline
\end{tabular}}
\end{center}
\caption{{System-level Kendall's tau correlation coefficients between reference-based and human evaluations under different reference summary settings. "Original" is the original heuristic reference from the dataset; others are generated by LLMs. References generated by LLMs improve the effectiveness of reference-based evaluations.}}
\label{tab:compare_kt_corr}
\end{table*}

%% file: ARR_version/04experiment.tex

\section{Benchmark Design}\label{sec:setup}

In this section, we first introduce the dataset for news summarization evaluation, followed by the SLM details, evaluation metrics, and reference summary generation.

\subsection{News Article Selection }

All experiments are conducted on the following four datasets: CNN/DM~\cite{cnndm}, XSum~\cite{xsum}, Newsroom~\cite{newsroom}, and BBC2024. The first three datasets have been widely used for verifying model text summarization~\cite{goyal2022news,google-2023-benchmarking,zhang2024benchmarking}. However, their older data may overlap with the training datasets of some models. To address this, we create BBC2024, featuring news articles from January to March 2024 on the BBC website~\footnote{\url{http://dracos.co.uk/made/bbc-news-archive/}}. Based on prior research, 500 samples per dataset are sufficient to differentiate the models~\cite{google-2023-benchmarking}. We randomly select 500 samples from each of the four test sets and construct 2000 samples in total. Each sample contains 500 to 1500 tokens according to the Qwen1.5-72B-Chat tokenizer.

\subsection{Model Selection}
We select 21 popular models with parameter sizes not exceeding 4 billion for benchmarking, including 19 SLMs and 2 models specifically designed for text summarization, Pegasus-Large and Brio, which have been fine-tuned on text summarization datasets, detailed in Table~\ref{tab:model_list}.

All models are sourced from the Hugging Face\footnote{\url{https://huggingface.co}} and tested based on the lm-evaluation-harness tool~\cite{eval-harness}. For general-purpose SLMs, we use the prompt from the huggingface hallucination leaderboard\footnote{\url{https://huggingface.co/spaces/hallucinations-leaderboard/leaderboard}} (Prompt 1 in Figure~\ref{fig:prompt}). To ensure reproducibility, we employ the greedy decoding strategy, generating until a newline character or <EOS> token is reached. Outputs are post-processed to remove prompt words and incomplete sentences. Given the 2048-token limit of some models, we only evaluate the zero-shot approach.

\begin{table}[]
\centering
\begin{adjustbox}{max width=1\columnwidth}
\begin{tabular}{lccc}
\hline
Model          & Parameters & \begin{tabular}[c]{@{}c@{}}Instruction \\ Tuning\end{tabular} & Reference                                                              \\ \hline
Brio $^{\ast}$          & 406M         & $\checkmark$                                                  & \cite{brio} \\
LiteLlama      & 460M         & $\times$                                                      & \cite{huggingface2024litelama}       \\
Qwen2-0.5B     & 500M         & $\times$                                                      & \cite{qwen}              \\
Qwen2-0.5B-Ins & 500M         & $\checkmark$                                                  & \cite{qwen}     \\
Bloom-560M          & 560M         & $\times$                                                      & \cite{bloom}        \\
Pegasus-Large$^{\ast}$  & 568M         & $\checkmark$                                                  & \cite{pegasus}         \\
TinyLlama      & 1.1B         & $\checkmark$                                                  & \cite{zhang2024tinyllama} \\
Llama3.2-1B    & 1.2B         & $\times$                                                      & \cite{llama3.2} \\
Llama3.2-1B-Ins& 1.2B         & $\checkmark$                                                  & \cite{llama3.2} \\
GPT-Neo-1.3B   & 1.3B         & $\times$                                                      & \cite{gpt-neo}      \\
Qwen2-1.5B     & 1.5B         & $\times$                                                      & \cite{qwen}              \\
Qwen2-1.5B-Ins & 1.5B         & $\checkmark$                                                  & \cite{qwen}     \\
InternLM2-1.8B & 1.8B         & $\times$                                                      & \cite{cai2024internlm2}     \\
InternLM2-1.8B-Chat & 1.8B    & $\checkmark$                                                  & \cite{cai2024internlm2}     \\
MiniCPM        & 2.4B         & $\checkmark$                                                  & \cite{hu2024minicpm}  \\
Gemma-1.1      & 2.5B         & $\checkmark$                                                  & \cite{team2024gemma}       \\
GPT-Neo-2.7B   & 2.7B         & $\times$                                                      & \cite{gpt-neo}       \\
Phi-2          & 2.7B         & $\times$                                                      & \cite{microsoft2023phi2}             \\
Llama3.2-3B    & 3.2B         & $\times$                                                      & \cite{llama3.2} \\
Llama3.2-3B-Ins& 3.2B         & $\checkmark$                                                  & \cite{llama3.2} \\
Phi-3-Mini     & 3.8B         & $\checkmark$                                                  & \cite{abdin2024phi3}   \\ \hline
\end{tabular}
\end{adjustbox}
\caption{\upshape{Models for news summarization benchmark. Models marked with $^{\ast}$ have been fine-tuned on text summarization datasets.} }
\label{tab:model_list}
\end{table}


\subsection{Evaluation Metric }

We evaluate the summary quality in relevance, coherence, factual consistency, and text compression using BertScore~\cite{bert-score}, HHEM-2.1-Open~\cite{HHEM-2.1-Open}, and summary length.


BertScore is a robust semantic similarity metric designed to assess the quality of the generated text and achieves the best correlation in Table~\ref{tab:compare_kt_corr}. It compares the embeddings of candidate and reference sentences through contextualized word representations from BERT~\cite{devlin-etal-2019-bert}. It can measure how well the summary captures key information (relevance) and maintains logical flow (coherence). We use the F1 score of BertScore\footnote{\url{https://huggingface.co/microsoft/deberta-xlarge-mnli}}, ranging from 0 to 100, where higher values indicate better quality.

HHEM-2.1-Open is a hallucination detection model which outperforms GPT-3.5-Turbo and even GPT-4. It can evaluate whether news summaries are factually consistent with the original article. It is based on a fine-tuned T5 model to flag hallucinations with a score. When the score falls below 0.5, it indicates a summary is inconsistent with the source. We report the percentage of summaries deemed factually consistent; higher values indicate better performance.

Since BertScore is insensitive to summary length, we also track the average summary length to assess text compression. With similar BertScore results, shorter summaries indicate better text compression ability.

\subsection{Reference Summary Generation}
To mitigate the impact of occasional low-quality reference summaries on evaluation results, as well as the tendency for models within the same series to score higher due to possible similar output content, we utilize two LLMs, Qwen1.5-72B-Chat and Llama2-70B-Chat, to generate two sets of reference summaries. We then average the scores during the SLM evaluation. We use Prompt 2 from Figure~\ref{fig:prompt} as the prompt template and apply a greedy strategy to generate summaries based on the vLLM inference framework~\cite{vllm}.

\section{Evaluation Results }
\label{sec:benchmark}


\subsection{Relevance and Coherence Evaluation}

\begin{table*}[]
\begin{center}
\resizebox{2\columnwidth}{!}{
\begin{tabular}{llccccccccr}
\hline
Score Range                  & Model         & \multicolumn{2}{c}{XSum} & \multicolumn{2}{c}{Newsroom} & \multicolumn{2}{c}{CNN/DM} & \multicolumn{2}{c}{BBC2024} & Average \\
                    &                    & Qwen1.5 & Llama2 & Qwen1.5 & Llama2 & Qwen1.5 & Llama2 & Qwen1.5 & Llama2 & \multicolumn{1}{l}{} \\ \hline
\multirow{4}{*}{$< 60$} & LiteLlama           & 38.68   & 39.24  & 38.02   & 38.70  & 40.58   & 41.38  & 38.05   & 38.86  & 39.19                \\
                    & Bloom-560M              & 48.23   & 48.43  & 42.73   & 43.28  & 49.70   & 50.04  & 47.56   & 48.01  & 47.25                \\
                    & GPT-Neo-1.3B            & 59.67   & 59.55  & 52.54   & 53.11  & 58.37   & 58.81  & 58.16   & 58.18  & 57.30                \\ 
                    & GPT-Neo-2.7B            & 60.11   & 60.15  & 53.65   & 54.14  & 59.15   & 59.97  & 59.29   & 59.50  & 58.25                \\ \hline
\multirow{10}{*}{$60 \sim 70$}
                    & Llama3.2-1B       &62.44    &62.33   &55.25    &55.86   &60.32    &60.67  &62.17&62.33      &60.17    \\
                    & Pegasus-Large      &61.39    &61.44   &61.26    &61.64   &61.94    &62.26  &63.12&63.47      &62.07    \\
                    & Llama3.2-3B       &66.14    &65.84   &59.93    &60.68   &64.19    &64.51  &66.62&66.57      &64.31    \\
                    & MiniCPM            & 65.59   & 65.77  & 62.48   & 63.13  & 66.82   & 67.51  & 65.54   & 65.96  & 65.35                \\
                    & TinyLlama          & 68.07   & 69.16  & 64.70   & 65.63  & 68.66   & 69.23  & 66.93   & 68.07  & 67.56                \\
                    & InternLM2-1.8B      & 68.64   & 69.20  & 64.27   & 65.00  & 68.81   & 69.49  & 67.88   & 68.74  & 67.75                \\
                    & InternLM2-1.8B-Chat & 67.77   & 68.82  & 64.99   & 65.57  & 68.78   & 69.38  & 68.32   & 69.62  & 67.91                \\
                    & Qwen2-0.5B-Ins & 69.30       & 70.01      & 65.85   & 66.65   &69.31   & 70.06 & 67.54 & 68.64        & 68.42   \\
                    & Qwen2-0.5B          & 69.57   & 70.71  & 65.94   & 66.92  & 69.50   & 70.48  & 68.23   & 69.67  & 68.88                \\
                    &Brio   & 70.67  & 70.71  &68.43  & 68.41   &69.86  &70.34  &70.39  & 70.21  & 69.88  \\ \hline
\multirow{7}{*}{$> 70$} & Phi-2              & 70.95   & 72.03  & 68.15   & 69.14  & 71.19   & 72.21  & 70.38   & 71.50  & 70.69                \\
                    & Gemma-1.1          & 70.80   & 72.27  & 69.42   & 70.30  & 71.88   & 72.83  & 70.60   & 71.67  & 71.22                \\
                    & Qwen2-1.5B          & 72.03   & 72.90  & 69.19   & 69.91  & 71.33   & 72.42  & 71.08   & 72.18  & 71.38                \\
                    & Qwen2-1.5B-Ins      & 72.21   & 73.06  & 69.49   & 69.86  & 71.64   & 71.95  & 71.01   & 71.97  & 71.40                \\
                    & Llama3.2-1B-Ins     & 72.24   & 73.27  & 70.54   & 70.98  & 72.43   & 73.61  & 71.68   & 72.55  & 72.16                \\
                    & Phi-3-Mini         & \textbf{74.67}   & {75.14}  & {72.42}   & {72.32}  & \textbf{74.86}   & {74.88}  & {74.08}   & {73.72}  & {74.01} \\  
                    & Llama3.2-3B-Ins     & 74.40   & \textbf{75.33}  & \textbf{72.81}   & \textbf{73.41}  & 74.54   & \textbf{75.29}  & \textbf{74.51}   & \textbf{74.95}  & \textbf{74.41}   \\ \hline
                  
\end{tabular}}
\end{center}
\caption{\upshape {BertScore of SLMs on four text summarization datasets. Qwen1.5-72B-Chat and Llama2-70B-Chat are used to generate reference summaries.}}
\label{tab:bertscore_res}
\end{table*}

\begin{figure}
    \centering
    \includegraphics[width=1\linewidth]{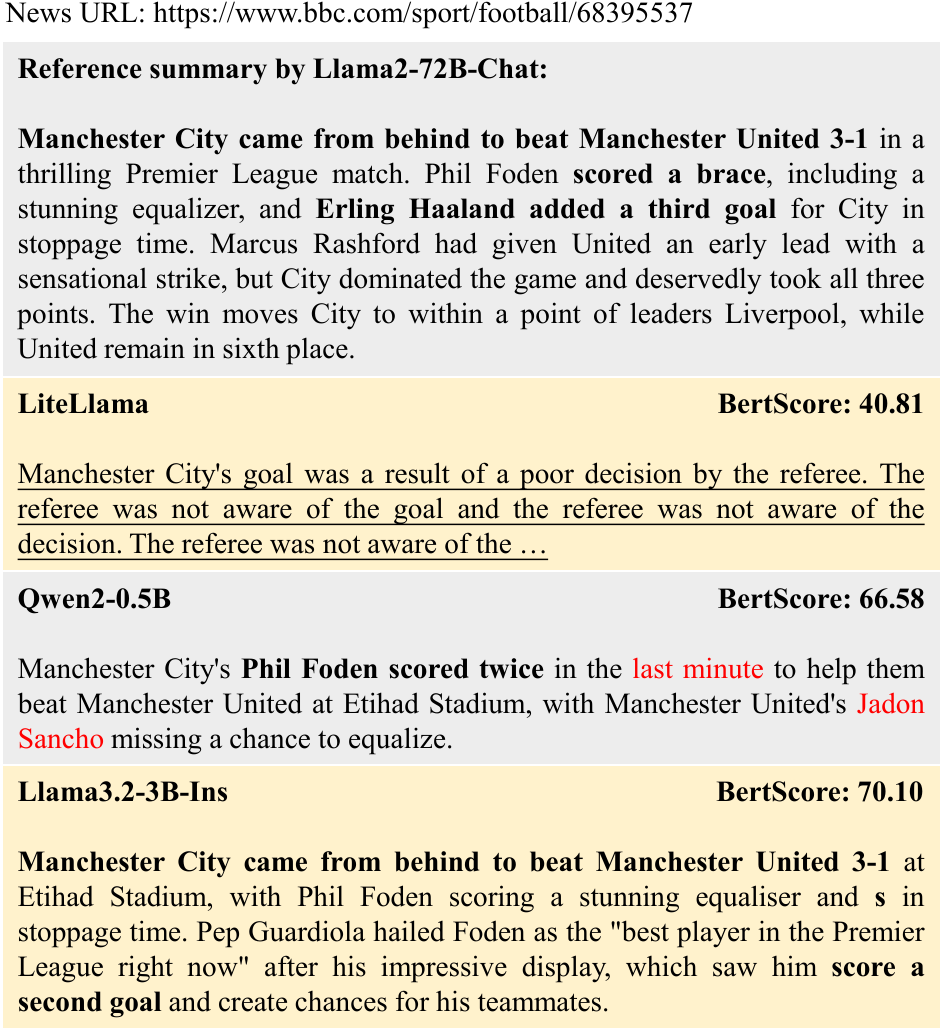}
    \caption{ Example summaries from SLMs and LLMs. The bold part is the same as the reference, the underline indicates irrelevant content, and the red indicates incorrect content. Summaries with BertScore above 70, such as those from Llama3.2-3B-Ins, demonstrate similar quality to LLMs. }
    \label{fig:bertscore_example}
\end{figure}

Table~\ref{tab:bertscore_res} shows the average BertScore results of SLMs. All models demonstrate consistent performance across datasets. Due to the significant score differences, we categorize the models into three approximate ranges, using two specialized summarization models, Pegasus-Large and Brio, as reference points.

The first range includes scores below 60. \textbf{Models scoring below 60 struggle to effectively summarize articles.} LiteLlama, Bloom, and GPT-Neo series fall into this range. In terms of relevance, these models often miss key points; in terms of coherence, these models sometimes produce repetitive outputs, leading to overly long summaries. Figure~\ref{fig:bertscore_example} shows an example from LiteLlama, it scores only 40.81, deviating significantly from the news.

The second range covers scores between 60 and 70. \textbf{Models in this range produce useful summaries but occasionally lack key points.} Models like TinyLlama and Qwen-0.5B series fall into this range. In relevance, these models generally relate to the original content but may omit crucial details. In coherence, the sentences can be well-organized. Fig~\ref{fig:bertscore_example} shows examples from Qwen2-0.5B. Although its summary is relevant and coherent to the news article, it omits the final score.

The third range includes scores above 70. \textbf{Models in this range produce summaries comparable to LLMs, with occasional inconsistencies.} They effectively capture key points and maintain coherent structure. Phi3-Mini and Llama3.2-3B-Ins stand out in this group. As shown in Figure~\ref{fig:bertscore_example}, Llama3.2-3B-Ins generates a concise, accurate summary, correctly capturing the match outcome even though the source news lacks a direct score description.

\subsection{Factual Consistency Evaluation}


During summary generation, SLMs may produce false information due to hallucinations. Therefore, we report the factual consistency rate in Table~\ref{tab:fact}. Models with high BertScore generally have better factual consistency as shown in Figre~\ref{fig:bertscore_example}. SLMs that perform well in fact consistency include Phi3-Mini (96.7\%) and Qwen2-1.5B-Ins (96.5\%), maintaining over 95\% consistency across all datasets. The traditional model, Pegasus-large, outperforms SLMs in fact consistency, which achieves an exceptional 99.9\% average. This is because it often copies sentences from the original text, ensuring alignment but sometimes missing key information.



\subsection{Summary Length Evaluation}

\begin{figure}
    \centering
    \includegraphics[width=1\linewidth]{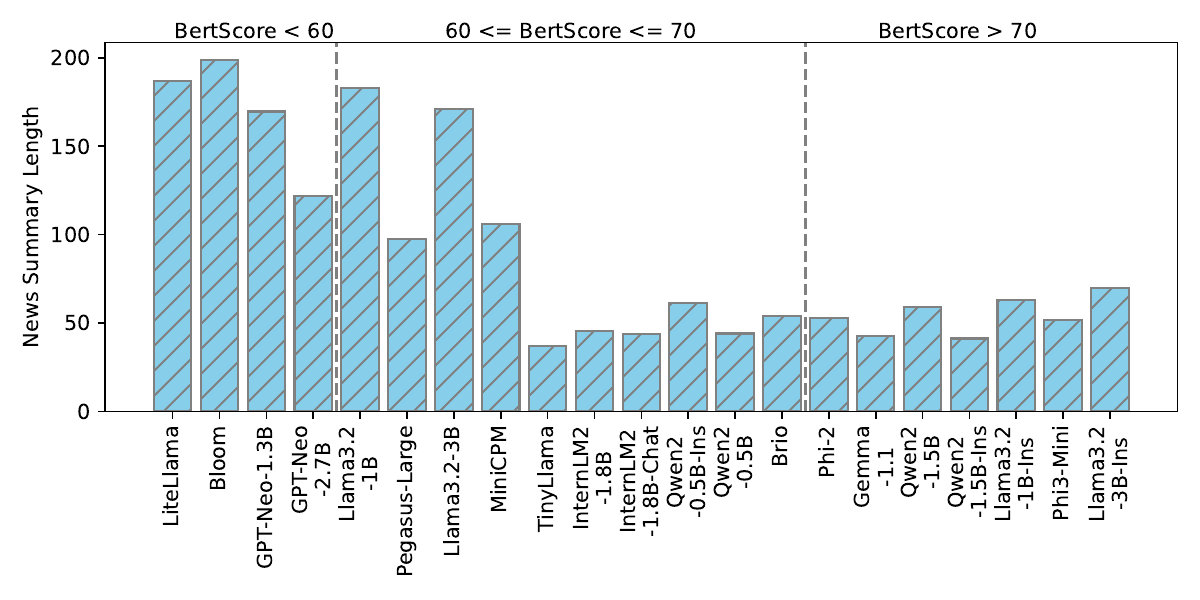}
    \caption{Average summary length comparison. SLMs with high BertScore generate 50-70 word summaries.}
    \label{fig:summary_length}
\end{figure}

Figure~\ref{fig:summary_length} shows the average summary lengths on 4 datasets produced by SLMs, which vary significantly across models. For scores below 60, summaries typically exceed 100 words due to redundant content and poor summarization. In the 60–70 range, length varies widely; e.g., the Llama3.2 series includes too many unnecessary details, while TinyLlama omits key information, leading to shorter summaries. Models with scores above 70 generate more consistent summaries, averaging around 50 to 70 words in length. For the top-performing models, Phi3-Mini and Llama3.2 series, Phi3-Mini generates more concise summaries while maintaining similar relevance.

\begin{table}[]
\begin{adjustbox}{max width=1\columnwidth}
\begin{tabular}{lccccc}
\hline
Model              & XSUM &Newsroom & CNN/DM & BBC2024 & Average \\ \hline
LiteLama           & 60.6 & 63.4    & 68.8   & 63.6    & 64.1    \\
Bloom-560M         & 83.6 & 77.4                          & 87.0   & 83.4    & 82.9    \\
GPT-Neo-1.3B       & 91.2 & 76.8                          & 92.4   & 92.0    & 88.1    \\
GPT-Neo-2.7B       & 83.2 & 67.0                          & 87.6   & 84.0    & 80.5    \\
Llama3.2-1B        & 97.6 & 85.8                         & 97.6  & 95.8      & 94.2    \\
Pegasus-Large      & 99.6 & 100.0                         & 100.0  & 100.0   & \textbf{99.9}    \\
Llama3.2-3B        & 98.8 & 85.6                         & 98.8    & 97.4    & 95.2    \\
MiniCPM            & 85.8 & 72.8                          & 93.6   & 89.6    & 85.5    \\
TinyLlama          & 95.4 & 92.4                          & 98.4   & 98.0    & 96.1    \\
InternLM2-1.8B     & 91.8 & 87.4                          & 96.0   & 94.8    & 92.5    \\
InternLM2-1.8B-Chat & 88.6 & 84.6                          & 93.8   & 90.8    & 89.5    \\
Qwen2-0.5B-Ins     & 87.2 & 86.0                          & 93.2   & 89.6    & 89.0    \\
Qwen2-0.5B         & 93.0 & 92.8                          & 96.8   & 95.2    & 94.5    \\
Brio               & 94.8 & 89.0                          & 97.8   & 95.4    & 94.3    \\
Phi2               & 95.2 & 93.4                          & 97.8   & 96.8    & 95.8    \\
Gemma-1.1          & 96.6 & 93.6                          & 97.8   & 94.6    & 95.7    \\
Qwen2-1.5B         & 95.6 & 96.2                          & 97.4   & 96.6    & 96.5    \\
Qwen2-1.5B-Ins     & 91.8 & 91.2                          & 94.4   & 92.4    & 92.5    \\
Llama3.2-1B-Ins    & 92.2 & 88.0                          & 95.4   & 94.0    & 92.4    \\
Phi3-Mini          & 96.0 & 94.8                          & 98.2   & 97.6    & \textbf{96.7}    \\ 
Llama3.2-3B-Ins    & 93.2 & 93.2                          & 96.6   & 95.6    & 94.7    \\ \hline
\end{tabular}
\end{adjustbox}
\caption{\upshape{Factual consistency rate (\%) on news datasets.}}
\label{tab:fact}
\end{table}


\subsection{Comparison with LLMs}



To further evaluate the SLM performance in news summarization, we compare the well-performing SLMs with LLMs, including ChatGLM3~\cite{glm2024chatglm}, Mistral-7B-Ins~\cite{jiang2023mistral}, Llam3-70B-Ins~\cite{llama3modelcard}, and Qwen2 series, all in instruction-tuned versions. As the models show similar performance across various reference summaries and datasets, we use Llama2-70B-Chat to generate references and present the average results on BBC2024 in Table~\ref{tab:vs_llm_bertscore}.

We highlight the top two metrics within each category. As can be seen, the summarization quality produced by the SLMs is on par with that of the LLMs. The difference in BertScore and factual consistency is minimal, with SLMs occasionally performing better. Notably, when BertScore scores are similar, smaller models tend to generate shorter summaries that facilitate quicker reading and comprehension of news outlines. 

\begin{table}[]
\centering
\begin{adjustbox}{max width=1\columnwidth}
\begin{tabular}{lccc}
\hline
Model & BertScore & \begin{tabular}[c]{@{}c@{}}Factual \\ consistency\end{tabular} & \begin{tabular}[c]{@{}c@{}}Summary\\ length\end{tabular} \\ \hline
Qwen2-1.5B-Ins  & 71.97 & 92.4 & \textbf{41} \\
Phi3-Mini       & 73.72 & \textbf{97.6} & \textbf{51} \\
Llama3.2-3B-Ins & \textbf{74.95} & 95.6 & 70 \\
ChatGLM3        & 73.41 & 96.0 & 59 \\
Mistral-7B-Ins  & 74.32 & 93.0 & 60 \\
Qwen2-7B-Ins    & 74.71 & 95.6 & 59 \\
Llama3-70B-Ins  & \textbf{76.06} & 95.8 & 71 \\
Qwen2-72B-Ins   & 73.78 & \textbf{98.4} & 86 \\ \hline
\end{tabular}
\end{adjustbox}
\caption{\upshape{Average metric comparison of SLMs and LLMs on BBC2024 dataset.}}
\label{tab:vs_llm_bertscore}
\end{table}

\begin{table}[]
\centering
\begin{adjustbox}{max width=1\columnwidth}
\begin{tabular}{lccc}
\hline
Model & \multicolumn{1}{r}{Prompt 1} & \multicolumn{1}{r}{Prompt 2} & \multicolumn{1}{r}{Prompt 3} \\ \hline
Qwen2-0.5B-Ins  & 68.64 & 60.31 & 62.46 \\
Qwen2-1.5B-Ins  & 71.97 & 69.74 & 72.30 \\
Llama3.2-1B-Ins & 72.55 & 72.16 & 67.53 \\
Phi3-Mini       & 73.72 & 73.27 & 73.83 \\
Llama3.2-3B-Ins & 74.95 & 74.94 & 75.12 \\ \hline
\end{tabular}
\end{adjustbox}
\caption{\upshape{BertScore comparison with different prompts.} }
\label{tab:diff_prompt_bertscore}
\end{table}



\subsection{Human Evaluation}

To further validate the effectiveness of our metrics, we performed a small-scale human evaluation, referencing existing studies~\cite{fabbri2021summeval}. Specifically, six highly educated annotators evaluated the relevance of news summaries generated by five SLMs based on the original news articles (scored on a 1-5 scale, with higher scores indicating greater relevance). We randomly selected 20 news instances from the BBC2024 dataset as our evaluation set. As shown in Table~\ref{tab:small_human_eva}, the average scores of each model demonstrate a strong alignment between BertScore and human evaluations, achieving a Kendall correlation coefficient of 1.

\begin{table*}[]
\begin{tabular}{lrrrrrrrr}
\hline
Model & \multicolumn{1}{l}{Ann. 1} & \multicolumn{1}{l}{Ann. 2} & \multicolumn{1}{l}{Ann. 3} & \multicolumn{1}{l}{Ann. 4} & \multicolumn{1}{l}{Ann. 5} & \multicolumn{1}{l}{Ann. 6} & \multicolumn{1}{l}{Average} & \multicolumn{1}{l}{BertScore} \\ \hline
Bloom-560M & 1.95 & 1.25 & 1.43 & 1.60 & 1.00 & 1.00 & 1.37 & 46.30 \\
Llama3.2-3B & 3.15 & 2.10 & 2.25 & 2.50 & 1.35 & 1.15 & 2.08 & 66.36 \\
Qwen2-0.5B & 4.05 & 2.15 & 3.28 & 2.70 & 3.60 & 3.25 & 3.17 & 69.85 \\
Phi3-Mini & 4.45 & 4.60 & 4.00 & 3.45 & 4.60 & 4.25 & 4.20 & 73.14 \\
Llama3.2-3B-Ins & 4.65 & 4.45 & 4.35 & 3.65 & 4.00 & 4.10 & 4.23 & 75.15 \\ \hline
\end{tabular}
\caption{Humance evaluation in relevance on 5 SLMs.}
\label{tab:small_human_eva}
\end{table*}

\subsection{Overall Evaluation and Model Selection}
SLMs exhibit considerable variability in text summarization performance, with larger and newer models generally showing stronger capabilities. Among them, Phi3-Mini and Llama3.2-3B-Ins perform the best, matching the 70B LLM in relevance, coherence, and factual consistency, while producing shorter summaries. All of these comparisons suggest that SLMs can effectively replace LLMs on edge devices for news summarization.

Although we evaluate 19 SLMs, there is considerable variation in their parameter sizes. To optimize deployment on edge devices, we provide model selection recommendations based on size. For models under 1B parameters, Brio and Qwen2-0.5B are the top choices overall. General-purpose language models in this range offer no clear advantage over those specialized in text summarization. For models between 1B and 2B parameters, Llama3.2-1B-Ins demonstrates a clear edge. For models above 2B parameters, Llama3.2-3B-Ins and Phi3-Mini outperform others across all criteria.

\section{Influencing Factor Analysis}
\label{sec:factor}

\subsection{Prompt Design}
Prompt engineering has demonstrated significant power in many tasks, helping to improve the output quality of LLMs~\cite{zhao2021calibrate,surveypromptengineering}. Therefore, we use different prompt templates shown in Figure~\ref{fig:prompt} to analyze the impact of prompt engineering on BertScore scores, factual consistency, and summary length. The instructions become more detailed from Prompt 1 to Prompt 3.

\begin{figure}
    \centering
    \includegraphics[width=1\linewidth]{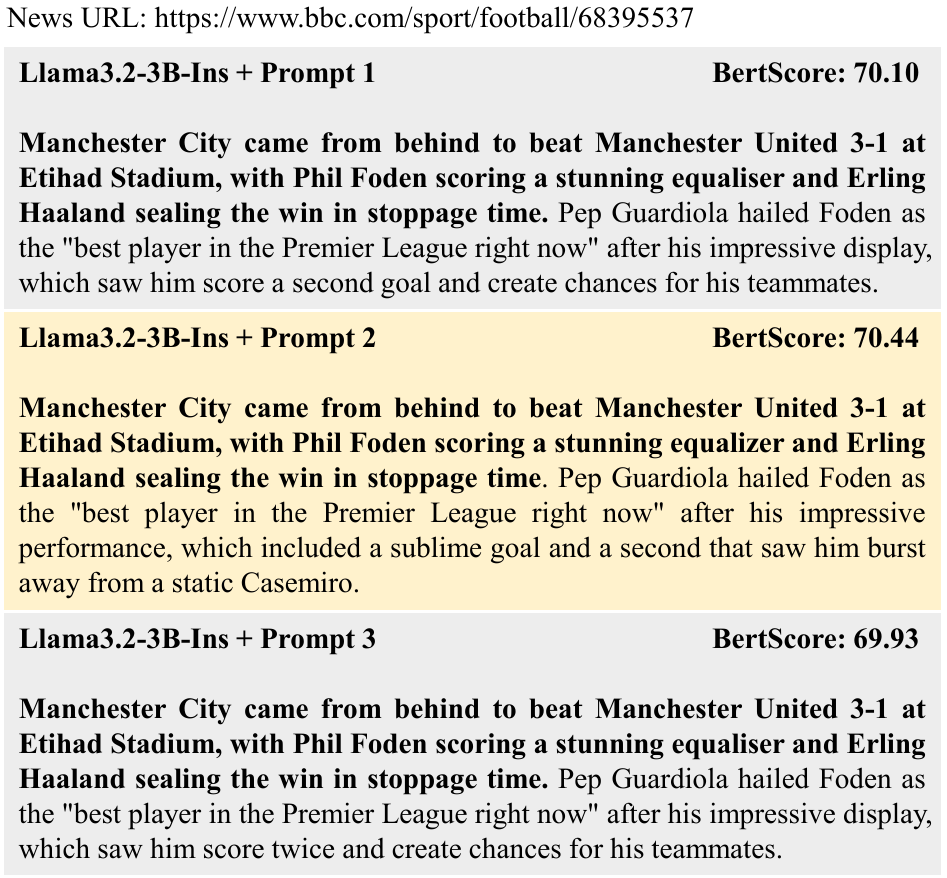}
    \caption{Examples summaries generated by Llama3.2-3B-Ins with different prompts. The bold parts show the same content, suggesting that the prompt design has a limited improvement in summary quality.}
    \label{fig:diff_prompt_example}
\end{figure}

Table~\ref{tab:diff_prompt_bertscore} compares the BertScore. The small score differences among prompts suggest that detailed descriptions do not significantly improve summary relevance and coherence and may even negatively affect some models. This suggests that \textbf{simple instructions are more suitable for SLMs, and overly complex prompts may degrade the performance}. For instance, the BertScore of Qwen2-0.5B-Ins drops significantly. The decline is likely due to prompt noise, where the model struggles to balance multiple instructions, resulting in less coherent summaries. Additionally, Figure~\ref{fig:diff_prompt_example} shows an example of Llama3.2-3B-Ins with different prompts. All summaries capture the key information, with only slight differences in details.

Table~\ref{tab:diff_prompt_fact} provides a comparison of factual consistency rates for various models using three different prompts. Many models, such as Qwen2-0.5B-Ins, exhibit a significant drop in factual consistency when using prompt 3. This further demonstrates that models with very small parameters are not well-suited for complex prompt design. The summary length variations across different prompts in Table~\ref{tab:diff_prompt_length} also indicate the limitations of prompt design. Some models fail to generate summaries of appropriate length given complex prompts.

\begin{table}[]
\centering
\begin{adjustbox}{max width=1\columnwidth}
\begin{tabular}{lccc}
\hline
Model & \multicolumn{1}{r}{Prompt 1} & \multicolumn{1}{r}{Prompt 2} & \multicolumn{1}{r}{Prompt 3} \\ \hline
Qwen2-0.5B-Ins  & 89.6 & 89.8  & 78.2 \\
Qwen2-1.5B-Ins  & 92.4 & 90.8  & 89.0 \\
Llama3.2-1B-Ins & 94.0 & 94.8  & 91.6 \\
Phi3-Mini       & 97.6 & 96.4  & 97.2 \\
Llama3.2-3B-Ins & 95.6 & 96.2  & 94.6 \\ \hline
\end{tabular}
\end{adjustbox}
\caption{\upshape{Factual consistency rate (\%)  comparison with different prompts.} }
\label{tab:diff_prompt_fact}
\end{table}

\begin{table}[]
\centering
\begin{adjustbox}{max width=1\columnwidth}
\begin{tabular}{lccc}
\hline
Model & \multicolumn{1}{r}{Prompt 1} & \multicolumn{1}{r}{Prompt 2} & \multicolumn{1}{r}{Prompt 3} \\ \hline
Qwen2-0.5B-Ins  & 65 & 150 & 151 \\
Qwen2-1.5B-Ins  & 41 & 99  & 53 \\
Llama3.2-1B-Ins & 63 & 60  & 58 \\
Phi3-Mini       & 51 & 57  & 65  \\
Llama3.2-3B-Ins & 70 & 67  & 61 \\ \hline
\end{tabular}
\end{adjustbox}
\caption{\upshape{Summary length with different prompts. All prompts require models to summarize in two sentences.} }
\label{tab:diff_prompt_length}
\end{table}

Overall, the improvement of prompt design on the quality of SLM summaries is limited. For SLMs that already perform well with simple prompts, they can capture the key points without the need for complex prompts. In contrast, SLMs that perform poorly may struggle with the logical relationships in complex prompts, leading to lengthy summaries. For practical applications, we recommend using simple and clear prompts when deploying SLMs for news summarization.


\subsection{Instruction Tuning}
Instruction tuning plays a crucial role in training LLMs by enhancing their ability to follow specific instructions. And some existing studies claim that instruction-tuned language models have stronger summarization capabilities~\cite{goyal2022news,zhang2024benchmarking}. However, in our extensive evaluations, with the exception of the Llama3.2 series, models perform very similarly before and after instruction tuning. In fact, models without instruction tuning often exhibit higher factual consistency. In the sample analysis, we find that the Llama3.2 models without instruction tuning include too many details, while instruction tuning helps it produce more general summaries. However, this trend is not observed in other models like Qwen2 and InternLM2 series, which is different from the conclusions of previous work~\cite{zhang2024benchmarking}. We leave the deep study of instruction tuning on summarization ability as a future research direction.

\section{Conclusion}
\label{sec:conclusion}
This paper presents the comprehensive evaluation of SLMs for news summarization, comparing 19 models across diverse datasets. Our results demonstrate that top-performing models like Phi3-Mini and Llama3.2-3B-Ins can match the performance of larger 70B LLMs while producing shorter, more concise summaries. These findings highlight the potential of SLMs for real-world applications, particularly in resource-constrained environments. Further exploration shows that simple prompts are more effective for SLM summarization, while instruction tuning provides inconsistent benefits, necessitating further research.


\section*{Limitations}

Although using LLM-generated summaries improved the reliability of reference-based summarization evaluation methods, it may introduce bias. For instance, if the LLM fails to accurately summarize the news, an SLM that produces similar content might receive an undeservedly high score. To mitigate this potential bias, we use summaries generated by multiple LLMs as references and calculate the average scores. Furthermore, although BertScore demonstrates a high level of consistency with human evaluations in coherence, it is not specifically designed to evaluate coherence. In future work, we plan to explore the use of more accurate metrics for coherence evaluation. Due to the input length limits of most SLMs, we only considered news articles with a maximum of 1,500 tokens. In the future, we will explore how to effectively use SLMs to summarize much longer news articles. Additionally, we do not consider quantized models in this study. In future work, we will also incorporate an analysis of the performance of quantized models into our evaluation.

%% file: ARR_version/05appendix.tex
\section{Challenges in Existing Summary Evaluation Method}
\label{sec:appendix}
In this section, we will supplement the detailed analysis of existing evaluation methods and explain why we use the reference-based evaluation method.

\subsection{Human evaluation}
\begin{figure}
    \centering
    \includegraphics[width=1\linewidth]{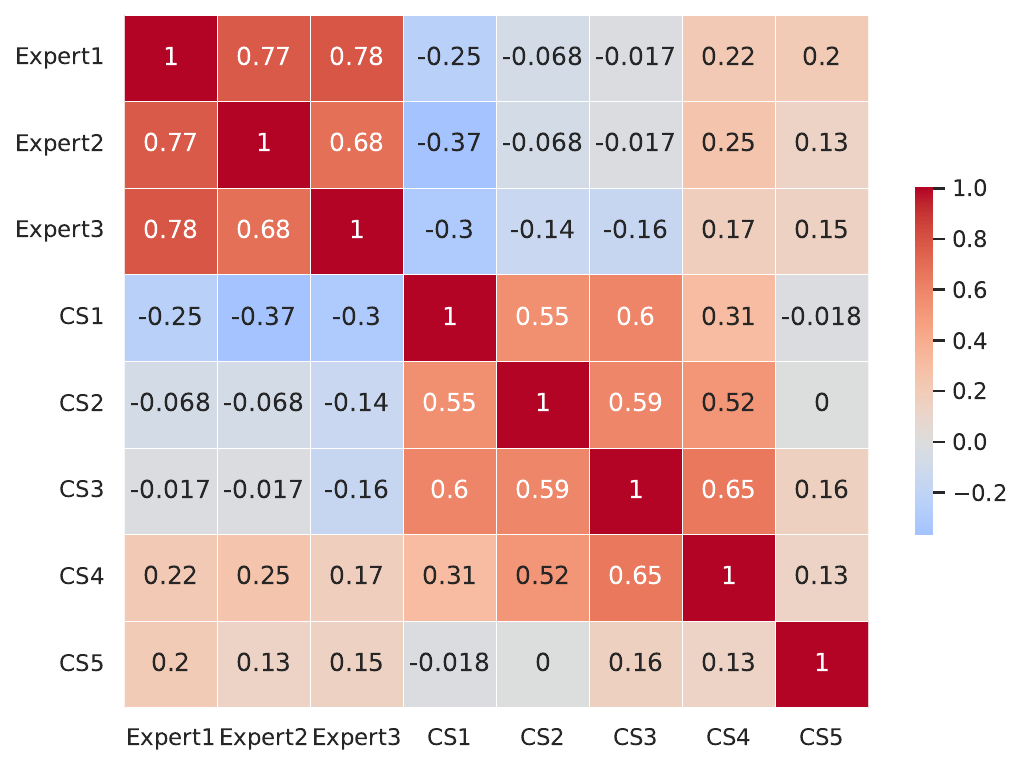}
    \caption{Pairwise system-level Kendall’s tau correlation on relevance for different annotators in SummEval benchmark. "CS" is the abbreviation for crowd-sourced. Darker colors indicate better consistency. }
    \label{fig:kd_heatmap_relevance}
\end{figure}

Human evaluation remains the most intuitive and effective method for text summarization evaluation~\cite{fabbri2021summeval,pu2023summarization,liu2023benchmarking,zhang2024benchmarking}. Annotators generally score generated summaries on aspects such as relevance and coherence (scale 1-5) based on the source text, and the final score is the average across all samples. Despite its reliability, human evaluation faces three main challenges for large-scale use: (1) Time-consuming—even evaluating 100 samples for each model can take weeks~\cite{zhang2024benchmarking}; (2) Low reproducibility—scores may vary due to different annotators; (3) High cost—only expert annotators are suitable, but hiring them is costly~\cite{fabbri2021summeval}. As shown in Figure~\ref{fig:kd_heatmap_relevance}, experts demonstrate significantly higher agreement than crowd-sourced annotators.

\subsection{Reference-based Evaluation}
\label{sec:reference eva}
Reference-based methods are among the earliest for evaluating text summarization~\cite{papineni-etal-2002-bleu}. They are more cost-effective and faster than human evaluation, using statistical metrics or small models to compute the similarity between generated and reference summaries. Higher similarity indicates better quality. There are two main categories: one measures textual overlap (e.g., ROUGE~\cite{lin-2004-rouge}, BLEU~\cite{papineni-etal-2002-bleu}); the other uses language models like BERT~\cite{devlin-etal-2019-bert} to compute semantic similarity (e.g., BertScore~\cite{bert-score}, BLEURT~\cite{sellam-etal-2020-bleurt}). However, the reference quality affects the evaluation reliability heavily~\cite{how_well,zhang2024benchmarking}. Heuristic methods used to create references in many datasets often lead to poor-quality summaries. For instance, XSum~\cite{xsum} uses only the first sentence of a news article as the reference, resulting in 54\% of summaries missing key information~\cite{how_well}. Moreover, since different datasets use varied reference generation methods, evaluation reliability fluctuates. Table~\ref{tab:compare_kt_corr} illustrates the impact of reference quality on reference-based methods. The correlation with human evaluation varies significantly across datasets when using the original reference for each dataset. It highlights the need for a unified, high-quality reference generation method.

\subsection{LLM evaluation}
Recent studies~\cite{goyal2022news,kendeer,gao2023human-like,liu2023benchmarking,wang2023chatgpt} have explored using LLMs for human-like summarization evaluation to achieve better alignment with human judgment. This approach involves guiding LLMs with prompts and providing both the summary and the original article for scoring. However, while LLM-based evaluation is reproducible and promising, it shares cost and efficiency challenges with human evaluation. Using commercial LLM APIs for large datasets can be expensive, and deploying open-source LLMs (e.g., 70B models) locally requires costly GPUs like the Nvidia A100 for reasonable speed. Additionally, studies~\cite{not-yet} indicate that LLMs may exhibit biases when evaluating different models, warranting further research to ensure fairness.

Overall, existing evaluation methods face issues related to financial cost, reproducibility, and effectiveness. Among them, reference-based methods are the most cost-effective and efficient. Therefore, we consider the LLM-augmented reference-based evaluation method as our metrics.



\section{Model Details}
In this section, we describe all benchmarked SLMs in our experiments.

\textbf{LiteLlama~\cite{huggingface2024litelama}:} LiteLlama is an open-source reproduction of Llama2~\cite{touvron2023llama2openfoundation}. It has 460M parameters and is trained on 1T tokens from the RedPajama dataset~\cite{together2023redpajama}.

\textbf{Bloom-560M~\cite{bloom}:} Bloom-560M is part of the BLOOM (BigScience Large Open-science Open-access Multilingual) family of models developed by the BigScience project. It has 560M parameters and is trained on 1.5T pre-processed text.

\textbf{TinyLlama~\cite{zhang2024tinyllama}:} TinyLlama has the same architecture and tokenizer as Llama 2. It has 1.1B parameters and is trained on 3T tokens. We use the chat finetuned version.

\textbf{GPT-Neo series~\cite{gpt-neo}:} GPT-Neo Series are open-source language models developed by EleutherAI to reproduce GPT-3 architecture. It has two versions with 1.3B and 2.7B parameters. They are trained on 380B and 420B tokens, respectively, on the Pile dataset~\cite{gao2020pile,biderman2022datasheet_pile}.

\textbf{Qwen2 series~\cite{qwen}:} Qwen2 comprises a series of language models pre-trained on multilingual datasets. And some models use instruction tuning to align with human preferences. We select models with no more than 7B parameters for benchmarking, including 0.5B, 0.5B-Ins, 1.5B, 1.5B-Ins, 7B, and 7B-Ins. Models with the "Ins" suffix indicate the instruction tuning version.

\textbf{InterLM2 series~\cite{cai2024internlm2}:} The InternLM2 series includes models with various parameter sizes. They all support long contexts of up to 200,000 characters. We select the 1.8B version and the 1.8B-Chat version for evaluation.

\textbf{Gemma-1.1~\cite{team2024gemma}:} Gemma-1.1 is developed by Google and is trained using a novel RLHF method. We select the 2B instruction tuning version for benchmarking.

\textbf{MiniCPM~\cite{hu2024minicpm}:} MiniCPM is an end-size language model with 2.7B parameters. It employs various post-training methods to align with human preferences. We select the supervised fine-tuning (SFT) version for benchmarking.

\textbf{Llama3.2 series~\cite{llama3.2}:} In the Llama 3.2 series, Meta introduced two SLMs, 1B and 3B, designed for edge devices. While Meta highlighted their capability to summarize social media messages, their performance in news summarization has yet to be evaluated.

\textbf{Phi series~\cite{microsoft2023phi2,abdin2024phi3}:} Phi series are some SLMs developed by Microsoft. They are trained on high-quality datasets. We select Phi-2(2.7B parameters) and Phi-3-Mini(3.8B parameters, 4K context length) for evaluation. 

\textbf{Brio~\cite{brio}: } Brio is one of the state-of-the-art specialized language models designed for text summarization, which leverages reinforcement learning to optimize the selection of sentences, ensuring high-quality, informative, and coherent summaries. It only has 406M parameters, and we select the version trained on the CNN/DM dataset.

\textbf{Pegasus-Large~\cite{pegasus}:} Pegasus-Large also is a specialized language model designed for abstractive text summarization, utilizing gap-sentence generation pre-training to achieve exceptional performance on summarization tasks. It has 568M parameters.